\documentclass[10pt, a4paper]{article}
\usepackage{lrec2022} 

\usepackage{multibib}
\newcites{languageresource}{Language Resources}
\usepackage{graphicx}
\usepackage{tabularx}
\usepackage{soul}
\usepackage{titlesec}
\titleformat{\section}{\normalfont\large\bfseries\center}{\thesection.}{1em}{}
\titleformat{\subsection}{\normalfont\SmallTitleFont\bfseries\raggedright}{\thesubsection.}{1em}{}
\titleformat{\subsubsection}{\normalfont\normalsize\bfseries\raggedright}{\thesubsubsection.}{1em}{}
\renewcommand\thesection{\arabic{section}}
\renewcommand\thesubsection{\thesection.\arabic{subsection}}
\renewcommand\thesubsubsection{\thesubsection.\arabic{subsubsection}}

\usepackage{times}
\usepackage{latexsym}

\usepackage[T1]{fontenc}
\usepackage[T5]{fontenc}

\usepackage[utf8]{inputenc}

\usepackage{microtype}

\usepackage{fancyvrb}

\usepackage{booktabs}
\usepackage{multirow}

\usepackage{subfig}
\usepackage{graphicx}

\usepackage{epstopdf}
\usepackage{hyperref}
\usepackage{xstring}

\usepackage{color}
\usepackage{float}
\usepackage{svg}
\usepackage{example}
\usepackage{examples}

\title{gaBERT --- an Irish Language Model}

\name{James Barry\textsuperscript{1}, Joachim Wagner\textsuperscript{2}, Lauren Cassidy\textsuperscript{1} \\
{\large
\bf{Alan Cowap\textsuperscript{3}}, \bf{Teresa Lynn\textsuperscript{1}}, \bf{Abigail Walsh\textsuperscript{1}}} \\
{\large 
\bf{Mícheál J. Ó Meachair\textsuperscript{4}}, \bf{Jennifer Foster\textsuperscript{2}}
}}

\address{$^{1,2,3}$School of Computing, Dublin City University,
$^{1}$ADAPT Centre \\
$^{3}$SFI Centre for Research Training in Machine Learning at Dublin City University \\
$^{4}$Fiontar \& Scoil na Gaeilge \\
$^{1}$ \texttt {\{firstname.lastname\}@adaptcentre.ie} \\
$^{2}$ \texttt {\{firstname.lastname\}@dcu.ie} \\
$^{3}$ \texttt {alan.cowap2@mail.dcu.ie} \\
$^{4}$ \texttt {micheal.omeachair@dcu.ie} \\
}

\pdfoutput=1

\definecolor{DarkRed}{RGB}{140,4,0}
\definecolor{DarkGreen}{RGB}{0,102,5}
\definecolor{DarkBlue}{RGB}{12,0,102}
\definecolor{Grey}{RGB}{144,144,144}

\newcommand{\TODO}[1]{}
\newcommand{\NOTE}[1]{}
\newcommand{\LATER}[1]{}

\begin{document}

\abstract{The BERT family of neural language models have 
become highly popular due to their ability to provide sequences of text with rich
context-sensitive token encodings which are able to generalise well to many NLP tasks.
We introduce gaBERT, a monolingual BERT model for the Irish language.
We compare our gaBERT model to multilingual BERT and the 
monolingual Irish WikiBERT,
and we show that gaBERT provides better representations for a downstream parsing task.
We also show how different filtering criteria, vocabulary size and the choice of subword tokenisation model
affect downstream performance.
We compare the results of fine-tuning a gaBERT model with an mBERT model for the task of identifying verbal multiword expressions, and show that the fine-tuned gaBERT model also performs better at this task.
We release gaBERT and related code to the community.
\\ \newline }

\maketitleabstract


\section{Introduction}


The technique of fine-tuning a self-supervised language model 
has become ubiquitous in Natural Language Processing (NLP)
because models trained in this way have advanced
evaluation scores
on many tasks~\cite{radford-etal-2018-improving,peters-etal-2018-deep,devlin-etal-2019-bert}.
Arguably the most popular architecture is BERT~\cite{devlin-etal-2019-bert} which uses stacks of
transformer blocks
to predict the identity of a masked token and to predict whether two sequences are contiguous.
It has spawned many variants
\cite{liu-etal-2019-roberta,lan-etal-2019-albert}
and 
much analysis
~\cite{jawahar-etal-2019-bert,chi-etal-2020-finding,rogers-etal-2020-primer}.
In this paper, we introduce gaBERT, a monolingual model of Irish.

%
 Although Irish is the first official language of the Republic of Ireland,
 only a minority, 1.5\% of the population 
 \cite{cso_census_2016},  
 use it in their everyday lives outside of the education system.
 As the less dominant language in a bilingual community, the availability of Irish language technology is important since it facilitates Irish speakers
 and learners
 to continue to use the language 
 in their increasingly digital daily lives. In terms of technological support however, Irish is a low-resourced language and significantly lacking in speech and language tools and resources \cite{lynn-ELE-2022}. 

 From a linguistic perspective, the Irish language is an inflected language, sharing linguistic features with other Celtic languages such as verb-subject-object (VSO) word order, initial mutation (lenition and eclipsis) and inflected prepositions. Inflection is common through suffixation, marking tense, number and person, while nouns are inflected for number and case. Nouns are either masculine or feminine in grammatical gender, which in turn influences declension-dependent inflections.
Its inflected nature has already been shown to impact data-driven NLP tools due to data sparsity \cite{lynn-etal-2013-working}, as has the frequent use of clefting (fronting), two forms of the verb `to be' and prevalence of variable and discontiguous multiword expressions.

 Building upon recent progress 
 in 
 data-driven
 Irish NLP~\cite{lynn-etal-2012-irish,lynn-etal-2015-minority,walsh-etal-2019-ilfhocail,cassidy2022}, we release gaBERT
 with the
 hope that it will contribute to
 preserving
 Irish as a living language in the digital age.

While there is evidence to suggest that dedicated monolingual models can be superior to a 
multilingual model for within-language downstream tasks \cite{devries2019bertje,virtanen-etal-2019-multilingual,farahani-etal-2020-parsbert}, other studies suggest that a multilingual model such as mBERT is a good choice for low-resourced languages \cite{wu-dredze-2020-languages,rust-etal-2020-how,chau-etal-2020-parsing}.
We compare gaBERT to mBERT
and to the monolingual Irish WikiBERT,
both using Wikipedia as the source of training data.
We base our comparison on the downstream task of
universal dependency (UD)
parsing, since we have labelled Irish data in the form of the Irish 
UD
Treebank~\cite{lynn-foster-2016-universal,mcguinness-etal-2020-annotating}.
We find that parsing accuracy improves when using gaBERT -- by
3.7 and 3.6 LAS points over mBERT 
and WikiBERT, respectively.
Continued pretraining of mBERT using the gaBERT training data results in
a recovery of 2 LAS points over the off-the-shelf version.
The benefit of the gaBERT training data is also shown in a manual analysis
which compares the models on their ability to predict a masked token,
as well as a Multiword Expression (MWE) identification task, where a token classification layer is trained to locate and classify verbal MWEs in text.

We detail our hyperparameter search for 
our final model,
where we consider the type of text filtering to apply,
the vocabulary size     
and tokenisation model.
We release our experiment
code  through
GitHub\footnote{\url{https://github.com/jbrry/Irish-BERT}} %
and our 
models
through the HuggingFace \cite{wolf-etal-2020-transformers} model repository.\footnote{\url{https://huggingface.co/DCU-NLP}}$^{,}$\footnote{We %
    also release gaELECTRA
    described in Appendix~\ref{sec:electra}.
}

\section{Data}\label{sec:data}
We use the following to train gaBERT:

\begin{itemize}
\item    \textbf{CoNLL17}: The Irish data from the CoNLL'17 raw text collection \cite{ginter-etal-2017-conll} released as part of the 2017 CoNLL Shared Task on UD Parsing \cite{zeman-etal-2017-conll}.
    
\item    \textbf{IMT}: A collection of Irish texts used in Irish machine translation research \cite{dowling-etal-2018-smt,dowling-etal-2020-human},
    including legal text, general administration and data crawled from public body websites. 
    
\item    \textbf{NCI}: The New Corpus for Ireland \cite{kilgarriff-etal-2006-efficient}, which 
    contains a wide range of texts in Irish, including fiction, news reports, informative texts and official documents. 
    
\item   \textbf{OSCAR}: The unshuffled Irish portion of the 2019 OSCAR corpus \cite{ortiz-suarez-etal-2019-asynchronous},
    a subset of CommonCrawl.
    
\item   \textbf{Paracrawl}: The Irish side of the \texttt{ga-en} bitext pair
    of ParaCrawl v7 
    \cite{banon-etal-2020-paracrawl}, which is a collection of parallel corpora crawled from multi-lingual websites.
    
\item    \textbf{Wikipedia}: Text from Irish Wikipedia,
    an online encyclopedia.\footnote{We use the articles from \url{https://dumps.wikimedia.org/gawiki/20210520/}
    }
    
\end{itemize}
The sentence and word counts
in each corpus are listed in Table~\ref{tab:corpus-sizes} after tokenisation and segmentation but before filtering described below.
See Appendix~\ref{app:data-licenses} for more information on the content of these corpora, including license information.
We apply corpus-specific pre-processing, sentence-segmentation and tokenisation, described in Appendix~\ref{sec:pp-corpora}.

\begin{table}
\begin{center}
\small
 \begin{tabular}{l r r r}
 \toprule
 Corpus & Num. Sents & Num. Tokens & Size (MB)  \\ [0.5ex] 
 \midrule
CoNLL17 & 1.7M & 24.7M & 138 \\
IMT & 1.4M & 22.6M & 124 \\
NCI & 1.6M & 33.5M & 174 \\
OSCAR & 0.8M & 16.2M & 89 \\
ParaCrawl & 3.1M & 67.5M & 380 \\
Wikipedia & 0.7M & 6.8M & 38 \\
\midrule
Overall & 9.3M & 171.3M & 943 \\    
\bottomrule
\end{tabular}
\end{center}
\caption{Sentence and word counts and plain text file size in megabytes for each corpus after tokenisation and segmentation but before applying sentence filtering.}
\label{tab:corpus-sizes}
\end{table}


\section{Experimental Setup}\label{sec:setup}
After initial corpus pre-processing,
all corpora
are merged and we use the WikiBERT pipeline \cite{pyysalo-etal-2020-wikibert}
to create pretraining data.
We experiment with four corpus filtering settings,
five vocabulary sizes
and three tokenisation models.

\subsection{Corpus Filtering}
\label{sec:filter}

The WikiBERT pipeline contains a number of filters 
which dictate whether a document should be kept.
As we are working with data sources where there may not be clear document boundaries,
or where there are no line breaks over a large number of sentences,
document-level filtering may be inadequate for such texts.
Consequently, we also experiment with using OpusFilter \cite{aulamo-etal-2020-opusfilter}, which filters individual sentences, 
thereby giving us the flexibility of filtering noisy sentences while not discarding full documents.

For each filter setting below, we train a BERT model on the data which remains after filtering:

\paragraph{No-filter}
All collected texts are included in the pre-training data.
\paragraph{Document-filter}
The default document-level filtering used in the WikiBERT pipeline.
\paragraph{OpusFilter-basic}
OpusFilter \cite{aulamo-etal-2020-opusfilter} with the following filters:
\begin{itemize}
    \setlength\itemsep{-0.5em}
    \item \texttt{LengthFilter}: Filter sentences containing more than 512 words.
    \item \texttt{LongWordFilter}: Filter sentences containing words longer than 40 characters.
    \item \texttt{HTMLTagFilter}: Filter sentences containing HTML tags.
    \item \texttt{PunctuationFilter}: Filter sentences which are over 60\% punctuation.
    \item \texttt{DigitsFilter}: Filter sentences which are over 60\% numeric symbols.
\end{itemize}
\paragraph{OpusFilter-basic-char-lang} The same filters are used as \textbf{OpusFilter-basic} but with additional character script and language ID filters:
\begin{itemize}
   \setlength\itemsep{-0.5em}
\item \texttt{CharacterScoreFilter}: All alphabetic characters in a sentence must be in Latin script.
\item \texttt{LanguageIDFilter}: The confidence scores from the language ID tools must be $>0.8$.
We use two language identification tools: \texttt{langid.py} \cite{lui-baldwin-2012-langid}
and CLD2.\footnote{\url{https://github.com/CLD2Owners/cld2}}
\end{itemize}


\subsection{Vocabulary Creation}\label{sec:vocab}
To create a model vocabulary,
we experiment with the SentencePiece \cite{kudo-richardson-2018-sentencepiece} and WordPiece tokenisers.\footnote{As BERT expects WordPiece tokenisation, a heuristic tool is used to map the SentencePiece vocabulary to WordPiece (\url{https://github.com/spyysalo/sent2wordpiece}).}
Using the model with highest median LAS from the filtering experiments,
we try vocabulary sizes of 15K, 20K, 30K, 40K and 50K.
We then train a WordPiece tokeniser, keeping the vocabulary size that works
best for the SentencePiece tokeniser.
We also train a BERT model using the union of the two vocabularies.

\subsection{BERT Pretraining Parameters}\label{sec:bert-train}

We use the original BERT implementation of \newcite{devlin-etal-2019-bert}.
For the development experiments,
we train our BERT model for 500K steps
with a sequence length of 128.
We use whole word masking
and
the default hyperparameters and model architecture of BERT\textsubscript{BASE} \cite{devlin-etal-2019-bert}.\footnote{We use a lower batch size of 32 in order to train on NVIDIA RTX 6000 GPUs with 24 GB RAM.}
Training for development runs of gaBERT took just under 48 hours on GPU.
While a seed for data preparation can be set
(we do not change the default 12345),
the BERT implementation
does not provide an option to set a seed for model
initialisation and we did not find code that
sets a seed for pretraining internally, suggesting initialisation is non-deterministic.

For the final gaBERT model, we train  for
900k steps with sequence length 128 and a further 100k steps
with sequence length 512.
We train on a TPU-v2-8 with 128GB of memory on Google Compute Engine\footnote{TPU access was kindly provided to us through the Google Research TPU Research Cloud.} and use a batch size of 128.
Training gaBERT on TPU for 1M steps took around 37.5 hours.




\section{Evaluation Measures}\label{sec:metrics}\label{sec:evaluation}

\paragraph{Dependency Parsing}\label{sec:metric:las}\label{sec:finetune}

The evaluation measure we use to make development decisions is
dependency parsing labelled attachment score (LAS).
To obtain this measure, we fine-tune a given BERT model
in the task of dependency parsing and measure LAS on the 
development set of the Irish-IDT treebank
in version 2.8 of UD.
We report the median of five fine-tuning runs with different random initialisation.
For the dependency parser,
we use a multitask model which uses
a graph-based parser with biaffine attention \cite{DBLP:journals/corr/DozatM16} 
as well as additional classifiers for predicting POS tags and morphological features.
Model hyperparameters are given in Appendix~\ref{sec:hyperparams-parser}.
We use the AllenNLP \cite{gardner-etal-2018-allennlp} library to develop our multitask model.

\paragraph{Cloze Test}
To compile a cloze task test set,
100 strings of Irish text (4--77 words each)
containing the pronouns ‘é’ (`him/it'), ‘í’ (`her/it') or ‘iad’ (`them') are selected
from Irish corpora and online publications.
One of these pronouns is masked in each string
for the cloze test.\footnote{All the
    masked tokens exist in the vocabularies of the candidate BERT models and are therefore
    possible predictions.
}

Following \newcite{ronnqvist-etal-2019-multilingual}, the models are evaluated on their ability to generate the original masked token, and a manual evaluation of the models is also performed wherein predictions are classified into the following exclusive categories:
\begin{itemize}
\setlength\itemsep{-00.2em}
\item\textbf{Match}: The predicted token fits the context grammatically and semantically. This may occur when the model predicts the original token or another token which also fits the context.
\item\textbf{Mismatch}
The predicted token is a valid Irish word but is unsuitable given the 
context.
\item\textbf{Copy}
The predicted token is an implausible repetition of another token in the context.
\item{\textbf{Gibberish}} The predicted token is not a valid Irish word. This might occur in the form of a subword or sequence of punctuation not forming a meaningful word.
\end{itemize}

\paragraph{MWE Identification task}
Multiword expressions (MWEs) pose a challenge in many tasks in NLP, including parsing. Treatment of MWEs can range between considering them as syntactically fixed words-with-spaces (ex: `out of', `Every cloud has a silver lining', `sooner or later'), to syntactically flexible constructions that display idiosyncratic behaviours (ex: `touch up', `life hack', `get something out of your system'). In addition to varying syntactic structures, MWEs also present issues of discontinuity, disambiguation, productivity, and can be more or less semantically opaque \cite{sag2002multiwordEA}. The task of automatically identifying multiword expressions (MWEs) has been explored in the series of shared tasks organised by the PARSEME network \cite{savary-etal-2017-parseme}, focusing on verbal MWEs, i.e. MWEs headed by a verb, as they pose a particular challenge in terms of automatic identification.

We design an experiment to compare the results of fine-tuning both a gaBERT model and an off-the-shelf mBERT model for the task of identifying MWEs in Irish text. We used the Irish portion of the PARSEME 1.2 shared task data \cite{walsh-etal-2020-annotating}, which has been manually annotated with six types of verbal MWEs. The annotations were converted to a modified version of \texttt{BIO} tagging, based on the work of \cite{schneider-etal-2014-discriminative}, and a linear layer for token classification was added for the task of identifying the correct label for each word.

\section{Results}\label{sec:results}
\subsection{Development Results}
%
\begin{table}[t]
\begin{center}
\small
\begin{tabular}{lrr}
\toprule
Filter      & Sentences & Tokens  \\ [0.5ex]
 \midrule
No-filter                  & 9.2M  &  171.3M \\
Document-filter            & 7.9M  &  161.0M \\
OpusFilter-basic           & 9.0M  &  170.8M \\
OpusFilter-basic-char-lang & 7.7M  &  161.2M \\
\bottomrule
\end{tabular}
\end{center}
\caption{The number of sentences and words which remain after applying the specific filter.}
\label{tab:filter-num-sentences}
\end{table}

%
\begin{figure}[b]
    \centering
    \includegraphics[width=8.5cm]{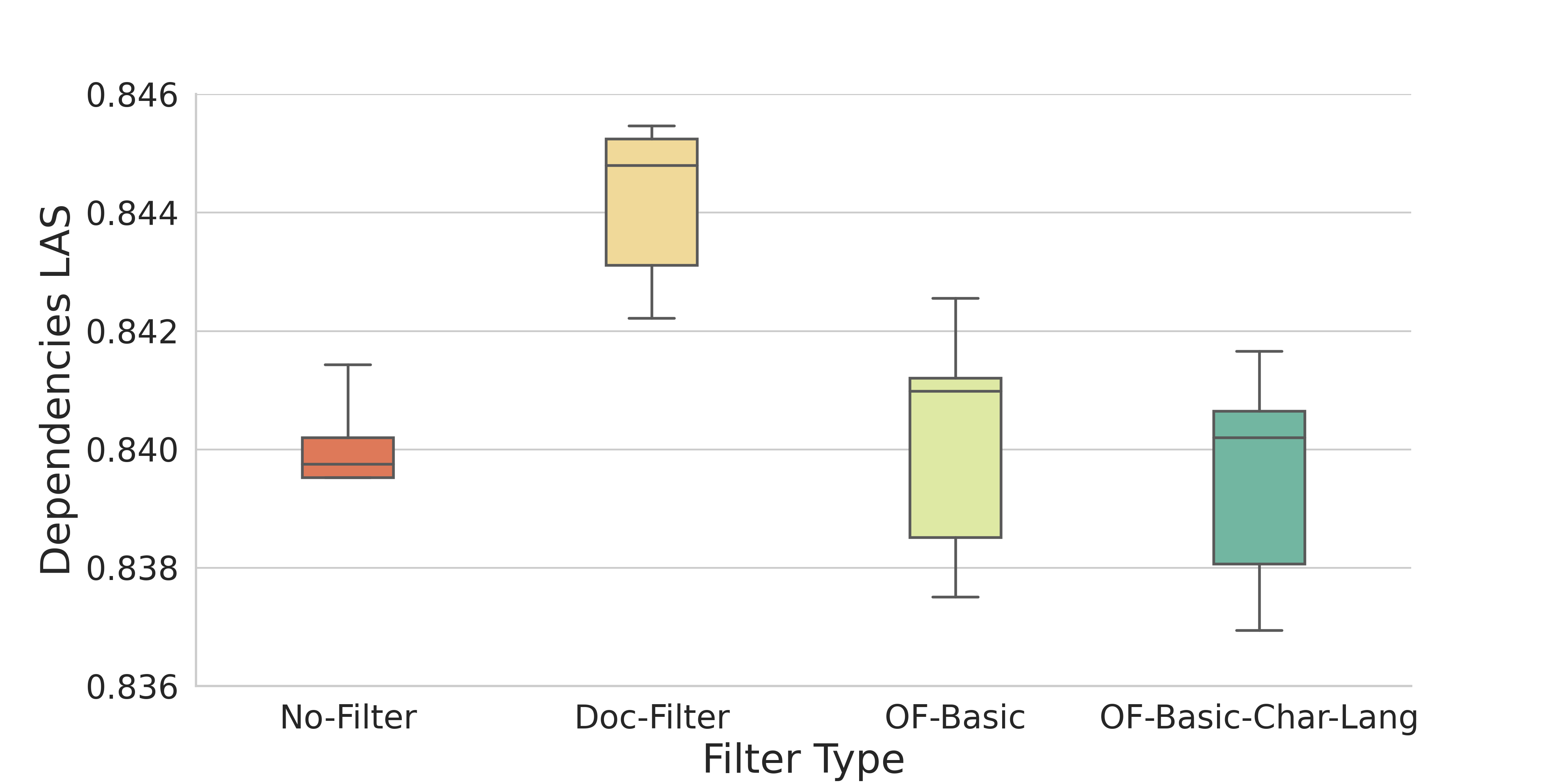}
    \caption{Dependency parsing LAS for each filter type.
    }
    \label{fig:filtering-LAS}
\end{figure}
%
\paragraph{Filter Settings}The overall number of sentences and words which remain after applying each filter are shown in Table~\ref{tab:filter-num-sentences}.
The results of training a dependency parser with the gaBERT model
produced by each setting are shown in
Fig.~\ref{fig:filtering-LAS}. 
\texttt{Document-Filter}
has the highest LAS score.
As the BERT model requires contiguous text for its next-sentence-prediction task,
filtering out full documents may be more appropriate than filtering individual sentences.
The two \texttt{OpusFilter} configurations perform marginally worse than the \texttt{Document-Filter}.
In the case of 
\texttt{OpusFilter-basic-char-lang},
the additional character script and language ID filters did not lead to a noticeable change in LAS.
Finally, \texttt{No-Filter} performs in the same range as the two \texttt{OpusFilter} configurations
but has the lowest median score,
suggesting that some level of filtering is beneficial.

\begin{figure}[t]
    \centering
    \includegraphics[width=8.5cm]{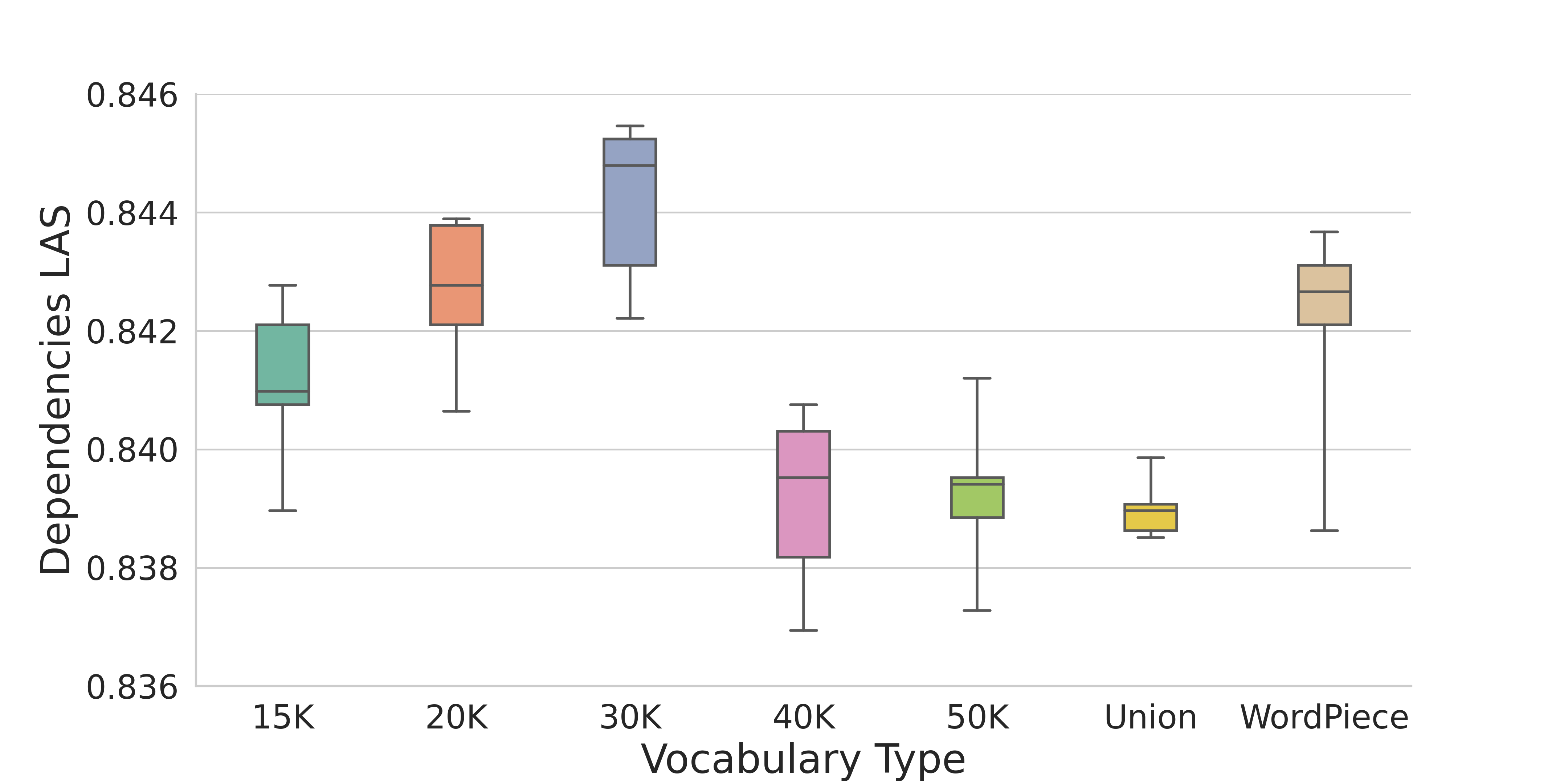}
    \caption{ Dependency parsing LAS for each vocabulary type.}
    \label{fig:vocab-LAS}
\end{figure}

\paragraph{Vocabulary Settings}
The results of the five runs testing different vocabulary sizes
are shown in 
Fig.~\ref{fig:vocab-LAS}.
A vocabulary size of 30K performs best for the SentencePiece tokeniser,
which outperforms the WordPiece tokeniser with the same vocabulary size.
The union of the two vocabularies results in 32,314 entries, and does not perform as well as the two vocabularies on their own.
A manual inspection of the two vocabularies showed that the WordPiece tokeniser created more entries consisting of foreign characters and emojis at the expense of Irish words/word-pieces, which may account for the lower performance of settings using this tokeniser.

\subsection{Model Comparison}
We compare our final gaBERT model with
off-the-shelf mBERT
and the monolingual Irish WikiBERT-ga model,
as well as 
an mBERT model
obtained with continued pre-training
on our corpora (mBERT-cp).\footnote{Since training the gaBERT model, other
    BERT models supporting Irish we found are
    BERTreach (\url{https://huggingface.co/jimregan/BERTreach}) 
    and
    LaBSE (\url{https://huggingface.co/setu4993/LaBSE}).
    BERTreach is a monolingual model trained
    on 47 million tokens. 
    LaBSE is a multilingual model trained to encode the meaning of sentences and covers 109 languages including Irish.
}

\paragraph{Dependency Parsing}

Table~\ref{tab:dev-results} shows the results
for dependency parsing.
The first row (No BERT)
is a baseline which does not use a pretrained BERT model
but uses a BiLSTM encoder operating over token and character-level features instead.
Using mBERT off-the-shelf results in a test set LAS of 80.3, an absolute improvement of 8.9 points over the baseline.
The WikiBERT-ga model
performs slightly better than mBERT.
By training  mBERT  for more steps on our corpora,
LAS can be improved by 2 points.
Our gaBERT model has the highest 
LAS of
84.

The last two rows compare gaBERT, on v2.5 of the treebank, with the system of \newcite{chau-etal-2020-parsing} 
who augment the mBERT vocabulary with the 99 most frequent Irish tokens and fine-tune on Irish Wikipedia.
The results are lower for both settings due to the fewer amount of trees in v2.5 of the treebank\footnote{v2.5 has only 858 trees compared to the 4,005 in v2.8.} and a manual effort to clean up some inconsistent annotations \cite{mcguinness-etal-2020-annotating}.
Our model outperforms this approach,
likely due to our inclusion of a wider variety of corpora as well as our dedicated Irish vocabulary.
 
\begin{table}
 \begin{center}
 \begin{tabular}{l r r r} 
 \toprule
     &      &  \multicolumn{2}{c}{LAS}   \\ [0.5ex]
  Model   & UD     &  Dev      &   Test     \\ [0.5ex]
 \midrule
  No BERT       &   2.8 &      73.4  &         71.4  \\
  mBERT       &   2.8 &      81.8  &         80.3  \\
  WikiBERT    &   2.8 &      81.9  &         80.4  \\
  mBERT-cp    &   2.8 &      84.3  &         82.3 \\
  gaBERT   & 2.8 &  \textbf{85.6} & \textbf{84.0} \\
  \midrule
  \newcite{chau-etal-2020-parsing} & 2.5 & - & 76.2 \\ 
  gaBERT & 2.5 & - & 77.5 \\ 
 \bottomrule
\end{tabular}
\end{center}
\caption{LAS in dependency parsing (UD v2.8) for selected models.
    Median of five fine-tuning runs.
    Scores are calculated using the official UD evaluation script (\textit{conll18\_ud\_eval.py}).
    }
    \label{tab:dev-results}
\end{table}
\paragraph{Cloze Test}
Table~\ref{tab:cloze-test-original-token-results} shows the accuracy of each model with regard to predicting the original masked token. mBERT-cp is the most accurate and gaBERT is
close behind.
\begin{table}
\begin{center}
 \begin{tabular}{l r}
 \toprule
 Model      & Original Token Prediction  \\ [0.5ex]
 \midrule
mBERT       &   16  \\
WikiBERT    &   53  \\
mBERT-cp    &  \bf 78  \\
gaBERT      &   75  \\
\bottomrule
\end{tabular}
\end{center}
\caption{The number of times the original masked token was predicted (100 test items).}
\label{tab:cloze-test-original-token-results}
\end{table}
%
%
\begin{table}
\begin{center}
 \begin{tabular}{l r r r r}
 \toprule
 Model      & Match & Mism. & Copy & Gib  \\ [0.5ex]
 \midrule
mBERT       &   41  &   42  &   4  &  13  \\
WikiBERT    &   62  &   31  &   1  &   6  \\
mBERT-cp    &   \bf 85  &   12  &   1  &   2  \\
gaBERT      &   83  &   14  &   2  &   1  \\
\bottomrule
\end{tabular}
\end{center}
\caption{The number of matches, mismatches, copies and gibberish predicted by each  model (100 test items).}
\label{tab:cloze-test-results}
\end{table}
Table~\ref{tab:cloze-test-results} shows the manual evaluation of the tokens generated by each model, accounting for plausible answers deviating from the original token
and separately reporting copying of content and production of gibberish.
These results echo those of the original masked token prediction evaluation in so far as they rank the models in the same order.

\begin{table*}
\small
\begin{tabular}{p{0.43\linewidth} | p{0.10\linewidth} | p{0.10\linewidth} | p{0.10\linewidth} | p{0.12\linewidth}}
 \toprule
\bf Context Cue & \bf Masked Word & \bf  Model & \bf Prediction & \bf Classification\\
\midrule
 \emph{Céard [MASK] na préamhacha raidiciúla sin?}\newline`What [MASK] those radical roots?' &
\emph{iad}\newline`them' &
mBERT-cp & \emph{faoi}\newline`about' & match\\
 \midrule
 \emph{Agus seo [MASK] an fhadhb mhór leis an bhfógra seo.}\newline`And this [MASK] the big problem with this advert.' & \emph{í}\newline`it' (fem.)
 &
WikiBERT & \emph{thaitin}\newline`liked' & mismatch\\
 \midrule
 \emph{Cheannaigh Seán leabhar agus léigh sé [MASK].}\newline`Seán bought a book and he read [MASK].' & \emph{é}\newline`it' (masc.)
 &
gaBERT & \emph{leabhar}\newline`a book' & copy\\
 \midrule
 \emph{Ní h[MASK] sin aidhm an chláir.}\newline`[MASK] is not the aim of the programme.' & \emph{\#\#é}\newline`it'(masc.)
 &
mBERT & \emph{-}\newline minus sign & gibberish\\
 \bottomrule
\end{tabular}

\caption{Examples of cloze test predictions and classifications.}
 \label{tab:cloze-examples}
\end{table*}

Table \ref{tab:cloze-examples} provides one example per classification category of masked token predictions generated by the language models during our cloze test evaluation.
In the \emph{match} example in Table \ref{tab:cloze-examples} , the original meaning (`What are those radical roots?') differs to the meaning of the resulting string (`What about those radical roots?') in which the masked token is replaced by the prediction of mBERT-cp. However, the latter construction is grammatically and semantically acceptable.
In the \emph{mismatch} example in Table \ref{tab:cloze-examples}, the predicted token is a valid Irish word, however the resulting generated text is nonsensical.
Though technically grammatical, the predicted token in the \emph{copy} example in Table \ref{tab:cloze-examples} results in a string with an unnatural repetition of a noun phrase where a pronoun would be highly preferable (`Seán bought a book and he read a book.').
In the \emph{gibberish} example in Table \ref{tab:cloze-examples}, the predicted token does not form a valid Irish word and the resulting sentence is ungrammatical.

In order to observe the effect that the amount of context provided has on the accuracy of the model, Table~\ref{tab:context-length-accuracy} shows the proportion of matches achieved by each language model when the results are segmented by the length of the context cues. 
\begin{table}
\begin{center}
 \begin{tabular}{l r r r}
 \toprule
 Model      & Short & Medium & Long \\ [0.5ex]
 \midrule
mBERT       & 20.69\% & 55.56\%  & 41.67\%  \\
wikibert    & 51.72\% & 58.33\%  & 74.29\%   \\
mBERT-cp    & 75.86\% & \bf83.33\%&\bf94.29\%\\
gaBERT      &\bf 79.31\%&\bf83.33\% & 85.71\% \\
gaELECTRA   &\bf 79.31\%& 77.78\% & 88.57\%  \\
\bottomrule
\end{tabular}
\end{center}

    \caption{Accuracy of language models segmented by length of context cue where short: 4--10 tokens, medium: 11--20 tokens, and long: 21--77 tokens.}
    \label{tab:context-length-accuracy}
\end{table}
All the models tested are least accurate when tested on the group of short context cues. All except mBERT achieved the highest accuracy on the group of long sentences.

A context cue may be considered easy or difficult based on:
\begin{itemize}
\setlength\itemsep{-00.2em}
\item Whether the tokens occur frequently in the training data \item The number of grammatical markers
\item The distance of the grammatical markers from the masked token
\end{itemize}
Two Irish language context cues which vary in terms of difficulty are exemplified below.

\begin{examples}
\item\label{ex:easy}\emph{\textbf{Bean}, agus \textbf{í} cromtha thar thralaí bia agus [MASK] ag ithe a \textbf{sáithe}.}\\
`A woman, bent over a food trolley while eating her fill.'\\
\end{examples}

 We can consider Example \ref{ex:easy} to be easy for the task of token prediction due to the following grammatical markers:
 \begin{itemize}
 \setlength\itemsep{-00.2em}
 \item `Bean' is a frequent feminine singular noun.
 \item `í' is a repetition of the feminine singular pronoun to be predicted.
 \item The lack of lenition on `sáithe' further indicates that the noun it refers to may not be masculine.
 \end{itemize}These grammatical markers indicate that the missing pronoun will be feminine and singular.\\
 
\begin{examples}
\item\label{ex:difficult}\emph{Seo \textbf{béile} aoibhinn fuirist nach dtógann ach timpeall leathuair a chloig chun [MASK] a ullmhú.}\\
`This is an easy, delicious meal that only takes about half an hour to prepare.'\\
\end{examples}
None of the language models tested predicted a plausible token for Example \ref{ex:difficult} This example is more challenging as the only grammatical marker is the feminine singular noun `béile' which is 11 tokens in distance from the masked token.

\paragraph{MWE Identification}

\begin{figure}[t]
    \centering
    \includegraphics[width=8.5cm]{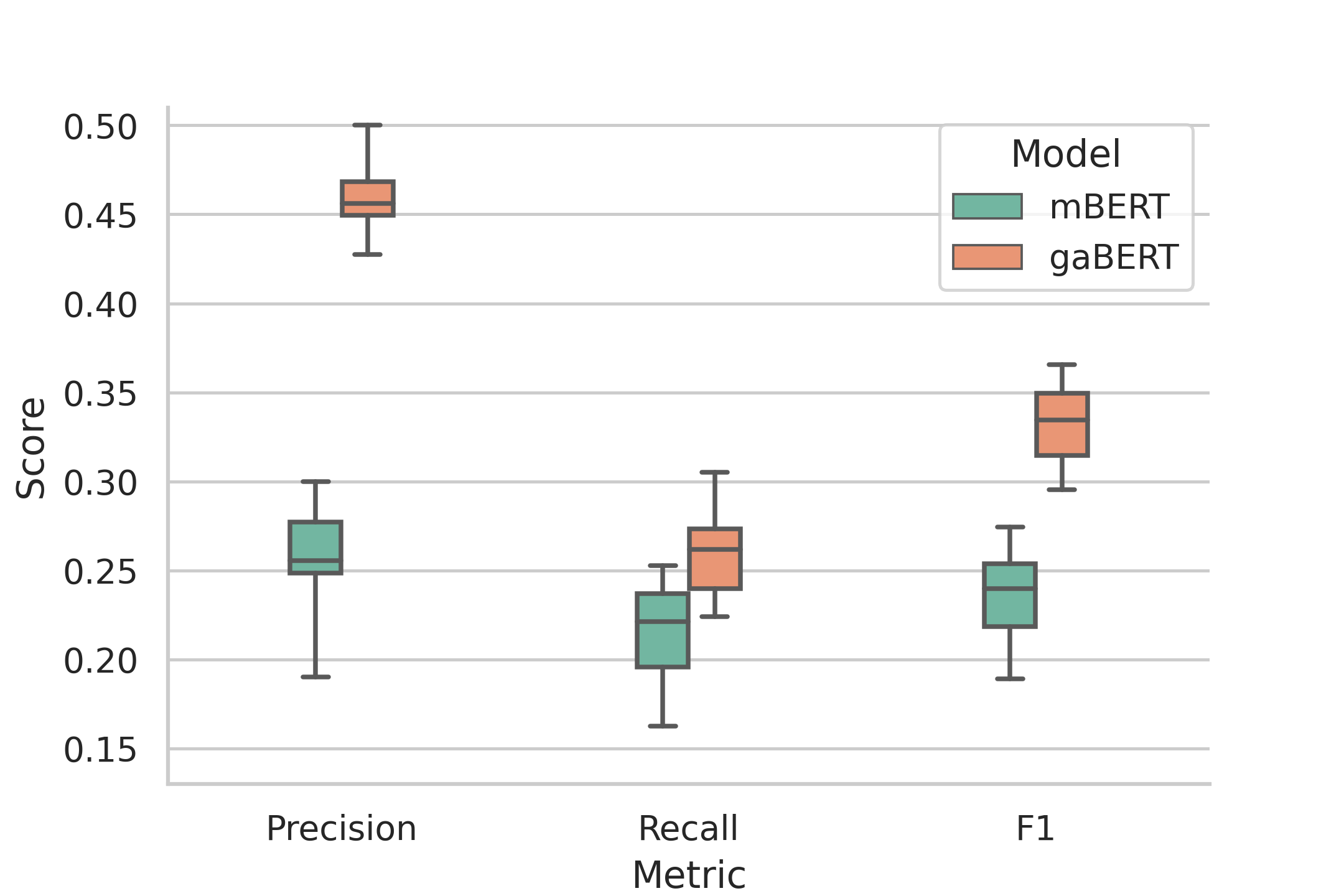}
    \caption{Verbal MWE Identification: Precision, Recall and F1 scores for each model across 20 random seed values}
    \label{fig:mwe-scores}
\end{figure}

MWE identification is a difficult task, and according to the system results of the most recent edition of the PARSEME shared task,\footnote{Full results: \url{http://multiword.sourceforge.net/sharedtaskresults2020/}} it appears to be particularly challenging for Irish, with
the majority of
systems performing most poorly on the Irish dataset. This may be due to the smaller size of the data, coupled with the relatively high number of MWE labels to classify \cite{walsh-etal-2020-annotating}. 
We attempt a series of fine-tuning experiments varying the learning rate, batch size and initial random seed, and found model performance is sensitive to changes in hyperparameters. Figure \ref{fig:mwe-scores} shows the results of training twenty models with different random seed values. It is evident that gaBERT outperforms mBERT in precision, recall and F1 scores on the test set.

Table \ref{tab:mwe-identification-results} records the Precision (P), Recall (R), and F1 scores for the best performing gaBERT and mBERT model found during the manual tuning of hyperparameters
(see Appendix~\ref{sec:hyperparams-mwe-identification} for details).
gaBERT performs better using these optimised parameters, particularly for precision scores, indicating that the gaBERT model tends to be correct more often when classifying MWEs than the mBERT model.

\begin{table}
\begin{center}
 \begin{tabular}{l r r r}
 \toprule
 Model  & P      & R      & F1  \\ [0.5ex]
 \midrule
 mBERT  & 0.342  & 0.245   & 0.285 \\
 gaBERT & \textbf{0.523}  & \textbf{0.361}   & \textbf{0.427} \\
\bottomrule
\end{tabular}
\end{center}
\caption{Verbal MWE Identification: (P)recision, (R)ecall and F1 scores of the best performing gaBERT and mBERT model}
\label{tab:mwe-identification-results}
\end{table}

In comparison to other systems submitted to the PARSEME shared task on the Irish data, both models perform well.
The best performing model for MWE identification had an F1 score of 0.306, which the gaBERT model exceeds by 0.121.\footnote{The results are not directly comparable, due to minor differences in calculating F1, so comparisons between our model and those systems submitted to the PARSEME shared task may be subject to slight variation when the same F1 calculation is used for both systems. Furthermore, emphasis in the most recent edition of the PARSEME shared task was on the identification of MWEs that had not been seen previously during the training phase. The highest ranking system for Irish actually had the second-highest F1 score for the task of global (seen and unseen) MWE identification, so we compare to the system with the highest F1 score for global MWE identification.} On a multilingual level, the averaged F1 score for overall MWE identification of the highest ranking system was 0.701, and even with the improved F1 score of the best performing gaBERT model, results for Irish are still below the best system for Hebrew (0.483), which was the language
where systems had the second-lowest performance.


\section{Friends of gaBERT}\label{sec:friends}

In subsequent experiments (see Appendix~\ref{sec:electra}-\ref{sec:full-model-results} for details), we look at variants of BERT, including RoBERTa~\cite{liu-etal-2019-roberta}.
The multilingual XLM-R\textsubscript{BASE} \cite{conneau-etal-2020-unsupervised} clearly outperforms both 
variants of mBERT  but underperforms  gaBERT.
We expect that the more diverse crawled data found in the XLM-R pretraining data makes it more competitive than mBERT.
We tried training a RoBERTa\textsubscript{BASE} model  but could only obtain  LAS scores comparable to off-the-shelf mBERT
and leave finding suitable hyperparameters to future work.  We train an ELECTRA model \cite{clark-etal-2020-electra}, which performs slightly below gaBERT but better than both mBERT models and the WikiBERT model.
As with gaBERT, this is likely due to the use of a dedicated Irish vocabulary which is absent in the multilingual models, and being exposed to more diverse data than Irish Wikipedia in the case of WikiBERT.

\section{Conclusions}\label{conclusion}
We release gaBERT, a BERT model trained on over
7.9M
Irish sentences
(containing approximately 161M words),
combining Irish language text from a variety of sources,
and evaluate it in dependency parsing, a pronoun cloze test task, and a MWE identification task,
showing improvements over three baselines,
multilingual BERT,
WikiBERT-ga and
XML-R\textsubscript{BASE}.

\section{Ethical Considerations}\label{sec:privacy}

No dataset is released with this paper, however most of the corpora are publicly available as described in 
Appendix~\ref{sec:data-availability}.
Furthermore, where an anonymised version of a dataset was available it was used.
We release the gaBERT 
language model
based on the BERT\textsubscript{BASE} \cite{devlin-etal-2019-bert}
autoencoder architecture. 
We note that an autoregressive architecture may be susceptible to training data extraction, and that larger language models may be more susceptible \cite{carlini-etal-2021-extracting}. However, gaBERT
is an
autoencoder architecture
and
a
smaller language model
which may help mitigate this potential vulnerability.

Possible harms of language model pre-trained on web-crawled text
have been widely discussed \cite{bender-etal-2021-on}. 
Since gaBERT uses CommonCrawl data, 
there is a risk that the gaBERT model may, for example,
produce unsuitable text outputs when used to generate text.
To mitigate this possibility we include the following caveat with the released code 
and 
model cards: 
\begin{quote}
    We note that some data used to pretrain gaBERT 
    was scraped from the web which potentially contains ethically problematic content (bias, hate, adult content, etc.).
    Consequently, downstream tasks/applications using gaBERT 
    should be thoroughly tested with respect to ethical considerations.
\end{quote}

We do not discuss in detail how gaBERT can be used in actual
use
cases as we expect the use of BERT-style models 
to be essential knowledge for NLP practitioners up-to-date with
current research.
There are many downstream tasks which can use gaBERT, 
including machine translation, educational applications, predictive text, search 
and games.
The authors hope gaBERT 
will contribute to the ongoing effort to preserve the Irish language as a living language in the technological age. Supporting a low-resourced language like Irish in a bilingual community will make it easier for Irish speakers, and those who wish to be Irish speakers, to use the language in practice.

Each use case or downstream application may 
rank the available
pre-trained language models differently in terms of suitability.
We urge NLP practitioners to compare available models such as those
tested in this paper in their application rather than relying
on results for a different task.

\section*{Acknowledgements}
This research is supported by Science Foundation Ireland (SFI) through the ADAPT Centre for
Digital Content Technology, which is funded under the SFI Research Centres Programme (Grant
13/RC/2106) and is co-funded under the European
Regional Development Fund.
This research is also supported through the SFI Frontiers for the
Future programme (19/FFP/6942) and SFI Centre for Research Training in
Machine Learning (18/CRT/6183), as well as by the Irish Government
Department of Culture, Heritage and the Gaeltacht
under the GaelTech Project.
We would like to thank Chris Larkin from
the TPU Research Cloud (TRC) for generously providing TPU access
and the anonymous reviewers for their helpful feedback and suggestions.
For the purpose of
Open Access, the authors have applied a CC BY
public copyright licence to any Author Accepted
Manuscript version arising from this submission.

\section*{References}  
\bibliography{main} 
\bibliographystyle{lrec2022-bib}

\appendix

\section{Data Licenses}\label{app:data-licenses}\label{sec:data-availability}

This Appendix provides specific details of the licence for each of the
datasets used in the experiments.

\subsection{CoNLL17}

The Irish annotated CoNLL17 corpus can be found here: 
\url{http://hdl.handle.net/11234/1-1989}
\cite{ginter-etal-2017-conll}.

The automatically generated annotations on the raw text data are available under the CC BY-SA-NC 4.0 licence.
Wikipedia texts are available under the CC BY-SA 3.0 licence.
Texts from Common Crawl are subject to Common Crawl Terms of Use, the full details of which can be found here: \url{https://commoncrawl.org/terms-of-use/full/}.

\subsection{IMT}

The Irish Machine Translation datasets contains text from the following sources:
\begin{itemize}
    \item Text crawled from the Citizen's Information website, contains Irish Public Sector Data licensed under a Creative Commons Attribution 4.0 International (CC BY 4.0) licence: \url{https://www.citizensinformation.ie/ga/}.
    \item Text crawled from Comhairle na Gaelscolaíochta website: \url{https://www.comhairle.org/gaeilge/}.
    \item Text crawled from the FÁS website (\url{http://www.fas.ie/}), accessed in 2017. The website has since been dissolved.
    \item Text crawled from the Galway County Council website: \url{http://www.galway.ie/ga/}.
    \item Text crawled from \url{https://www.gov.ie/ga/}, the central portal for government services and information.
    \item Text crawled from articles on the Irish Times website.
    \item Text crawled from the Kerry County Council website: \url{https://ciarrai.ie/}.
    \item Text crawled from the Oideas Gael website: \url{http://www.oideas-gael.com/ga/}.
    \item Text crawled from articles generated by Teagasc, available under PSI licence.
    \item Text generated by Conradh na Gaeilge, shared with us for research purposes. 
    \item The Irish text from a parallel English–Irish corpus of legal texts from the Department of Justice. This dataset is available for reuse on the ELRC-SHARE repository under a PSI license: \url{https://elrc-share.eu}
    \item Text from the Directorate-General for Translation (DGT), available for download from the European Commission website. Reuse of the texts are subject to Terms of Use, found on the website:  \url{https://ec.europa.eu/jrc/en/language-technologies/dgt-translation-memory}. 
    \item Text reports and notices generated by Dublin City Council, shared with us for research purposes.
    \item Text uploaded to ELRC-share via the National Relay Station, shared with us for research purposes.
    \item Text reports and reference files generated by the Language Commissioner, available on ELRC-share under PSI license: \url{https://elrc-share.eu/}.
    \item Text generated by the magazine Nós, shared with us for research purposes.
    \item Irish texts available for download on OPUS, under various licenses: \url{https://opus.nlpl.eu/}
    \item Text generated from in-house translation provided by the then titled Department of Culture, Heritage and Gaeltacht (DCHG), provided for research purposes. The anonymised dataset is available on ELRC-share, under a CC-BY 4.0 license: \url{https://elrc-share.eu/}.
    \item Text reports created by Údarás na Gaeilge, uploaded to ELRC-share available under PSI license: \url{https://elrc-share.eu/}.
    \item Text generated by the University Times, shared with us for research purposes.
\end{itemize}

\subsection{NCI}
The corpus is compiled and owned by Foras na Gaeilge and is provided to us for research purposes.

\subsection{OSCAR}
The unshuffled version of the Irish part of the
2019  
OSCAR corpus was provided to us by the authors for research purposes.

\subsection{ParaCrawl}
Text from ParaCrawl v7, available here: \url{https://www.paracrawl.eu/v7}.
The texts themselves are not owned by ParaCrawl, 
the actual packaging of these parallel data are under the Creative Commons CC0 licence ("no rights reserved").

\subsection{Wikipedia}
The texts used are available under a CC BY-SA 3.0 licence and/or a GNU Free Documentation License.

\section{Corpus Pre-processing}\label{sec:preproc-appendix}\label{sec:pp-corpora}

This appendix provides specific details on corpus pre-processing, and the OpusFilter filters used.

\paragraph{CoNLL17}
The CoNLL17 corpus is already tokenised, 
as it is provided in CoNLL-U format, which we convert to
one-sentence-per-line
tokenised plain text.

\paragraph{IMT, OSCAR and ParaCrawl}
The text files from the IMT, OSCAR and ParaCrawl contain raw sentences requiring tokenisation.
We describe the tokenisation process for these corpora in
Appendix~\ref{sec:tokenisation}.

\paragraph{Wikipedia}
For the Wikipedia articles, the Irish Wikipedia dump is downloaded and
the WikiExtractor tool\footnote{\url{https://github.com/attardi/wikiextractor}}
is then used to extract 
plain text.
Article headers are included in the extracted text files.
Once the articles have been converted to plain text,
they are tokenised using the tokeniser described in
Appendix~\ref{sec:tokenisation}.

\paragraph{NCI}
As many of the NCI segments
marked up with \texttt{$\langle$s$\rangle$} tags contain multiple sentences, we further split these segments with
heuristics described in Appendix~\ref{sec:pp-corpora-sbd}.

\subsection{Tokenisation and Segmentation}
\label{sec:tokenisation}

Raw texts from the IMT, OSCAR, ParaCrawl and Wikipedia corpora are tokenised and segmented with UDPipe \cite{straka-strakova-2017-tokenizing} trained on a combination of the Irish-IDT and English-EWT corpora from version 2.7 of the Universal Dependencies (UD) treebanks
\cite{11234/1-3424}. 
We include the English-EWT treebank in the training data to expose the tokeniser to more incidences of punctuation symbols which are prevalent in our pre-training data. 
This also comes with the benefit of supporting the tokenisation of code-mixed data.
We upsample the Irish-IDT treebank by ten times to offset the larger English-EWT treebank size.
This tokeniser is applied to all corpora apart from the NCI,
which is already tokenised by \newcite{kilgarriff-etal-2006-efficient},
and the CoNLL17 corpus as this corpus is already tokenised in CoNLL-U format.

\subsection{NCI}
\label{sec:pp-corpora-nci}

Foras na Gaeilge provided us with a \texttt{.vert}
file\footnote{MD5 7be5c0e9bc473fb83af13541b1cd8d20}
containing 33,088,532 tokens in 3,485 documents.
We extract the raw text from the first tab-separated column and
carry out the following conversions
(number of events):
\begin{itemize}

    \item Replace \texttt{\&quot;} with a neutral double quote
          (4408).
    
    \item Replace the standard xml/html entities quot, lt, gt and amp
          tokenised into three tokens, 
          with the appropriate characters
          (128).

    \item Replace the numeric html entities
          38, 60, 147, 148, 205, 218, 225, 233, 237, 243 and 250,
          again spanning three tokens,
          with the appropriate Unicode characters
          (3679).
          
    \item Repeat from the start until the text does not change.
    
\end{itemize}

We do not modify the seven occurrences of \texttt{\textbackslash x\textbackslash x13} as
it is not clear from their contexts how they should be replaced.
After pre-processing and treating all whitespace as
token separators, e.g. in the NCI token ``go leor'',
we obtain
33,472,496 tokens from the NCI.

\subsection{Sentence Boundary Detection}
\label{sec:pp-corpora-sbd}

Many of the NCI segments
marked up with
\texttt{$\langle$s$\rangle$} 
tags contain multiple sentences. We treat each
segment boundary as a sentence boundary and further split segments into
sentences recursively, finding the best split point
(among candidate split points after ``.'', ``?'' and ``!'' tokens)
according to the following heuristics, splitting the segment into two halves and applying the same procedure to each half until no suitable split point is found.
\begin{itemize}
    \item Reject if the left half contains no letters and is short.
          This includes cases where the left half is only a decimal number such as in enumerations.
    \item Reject if the right half has no letters and is short
          or is an ellipsis.
    \item Reject if the right half's first letter, skipping
          alphabetic and Roman enumerations in round brackets,
          is lowercase.
    \item Reject if the left half only contains a Roman number
          (in addition to the full-stop).
          
    \item Reject if inside round, square, curly or angle brackets
          and the brackets are not far away from the candidate split point.
\end{itemize}

For full-stop only:
\begin{itemize}
          
    \item Reject after ``DR'', ``Prof'' and ``nDr''.
          
    \item Reject after ``No'', ``Vol'' and ``Iml'' if
          followed by a decimal number.
          
\end{itemize}

Additional candidate split points are added with the following heuristics.
Furthermore, when we need to choose between multiple candidate split points that pass the above tests, we try to keep the lengths of the
halves (in characters)
similar but also factor in the preferences in the heuristics below.
\begin{itemize}
          
    \item If sentence-ending punctuation is followed by two quote
          tokens we
          also consider splitting between the quotes and prefer this
          split point if not rejected by above rules.
          
    \item If sentence-ending punctuation is followed by a closing
          bracket we
          also consider splitting after the closing bracket
          and prefer this
          split point if not rejected by above rules.
          
    \item If a question mark is followed by more question marks
          we also consider splitting after the end of the sequence of
          question marks and prefer this
          split point if not rejected by above rules.
          
    \item If a exclamation mark is followed by more exclamation marks
          we also consider splitting after the end of the sequence of
          exclamations marks and prefer this
          split point if not rejected by above rules.
    
    \item If a full-stop is the first full-stop in the overall
          segment, the preceding token is ``1'', there are
          more tokens before this ``1'' and the token directly
          before ``1'' is not a comma or semi-colon
          we assume that this is an enumeration following a
          heading and prefer splitting before the ``1''.

    \item Splitting after a full-stop following decimal numbers
          in all other cases is dispreferred, giving the largest
          penalty to small numbers as these are most likely to
          be part of enumerations.
          An exception is ``Airteagal'' followed by a token
          ending with a full-stop, a number, a full-stop,
          another number and another full-stop.
          Here, we implemented a preference for splitting
          after the first separated full-stop, assuming the
          last number is part of an enumeration.
    
\end{itemize}

\section{Hyperparameters used in the Multitask Parser and MWE Identification Task}
\label{sec:hyperparams-parser-mwe-identification}
This appendix provides specific details and hyperparameters for the multitask parser and MWE identification model.

\subsection{Multitask Parser}
\label{sec:hyperparams-parser}
The hyperparameters of the multitask parser are given in Table~\ref{table:hyper-params}.
For the tagging tasks,
the output of the Transformer is 
first projected through a task-specific Feedforward network and then passed to a classification layer.
For dependency parsing,
the projected representations from the tagging modules are concatenated to the output of the Transformer before being passed to the parsing module.

\begin{table}[h]
\begin{center}
\begin{tabular}{ll}
\multicolumn{2}{c}{\bf Multitask Parser Details } \\
\toprule
\noalign{\vskip 2mm}
\multicolumn{2}{c}{\bf Encoder } \\
Word-piece embedding size & 768 \\
Word-piece type & average \\
\noalign{\vskip 2mm}
\multicolumn{2}{c}{\bf Tagger (UPOS/XPOS/Feats)} \\
MLP size     & 200 \\
Dropout MLP        & 0.33 \\
Nonlinear act. (MLP) & ELU \\
\noalign{\vskip 2mm}
\noalign{\vskip 2mm}
\multicolumn{2}{c}{\bf Parser } \\
Arc MLP size     & 500 \\
Label MLP size & 100 \\
Dropout LSTMs      & 0.33 \\
Dropout MLP        & 0.33 \\
Dropout embeddings & 0.33 \\
Nonlinear act. (MLP) & ELU \\
\noalign{\vskip 2mm}
\multicolumn{2}{c}{\bf Optimiser and Training Details } \\
Optimizer          & AdamW \\
Learning rate      & 3e-4 \\
beta1           & 0.9 \\
beta2              & 0.999 \\
Num. epochs              & 50 \\
Patience              & 10 \\
Batch size              & 16 \\
\bottomrule
\end{tabular}
\end{center}
\caption{\label{table:hyper-params} Chosen hyperparameters for the multitask parser and tagger.}
\end{table}

\subsection{MWE Identification}
\label{sec:hyperparams-mwe-identification}
For the task of automatically identifying MWEs, the best performing models were found using a learning rate of 2e-5, and a random seed of 10. We trained the models for 20 epochs each. Using a batch size of 5, we found the best performing mBERT model, while the best performing gaBERT model used a batch size of 1. We fine-tuned each model on all layers.

\section{gaELECTRA Model}\label{sec:electra}
In addition to the gaBERT model of the main paper, we release
gaELECTRA,
an ELECTRA model \cite{clark-etal-2020-electra}
trained on the same data as gaBERT.
ELECTRA replaces the MLM pre-training objective of BERT
with a binary classification task
discriminating between authentic tokens and 
alternative tokens generated by a smaller model
for higher training efficiency.
We use the default settings of the ``Base'' configuration of the official
implementation\footnote{\url{https://github.com/google-research/electra}} and train on a TPU-v3-8.
As with BERT, we train for 1M steps and evaluate every 100k steps.
However, we train on more data per step as the batch size is increased from 128 to
256 and a sequence length of 512 is used throughout.

\begin{figure}[htp]
    \centering
    \includegraphics[width=8.5cm]{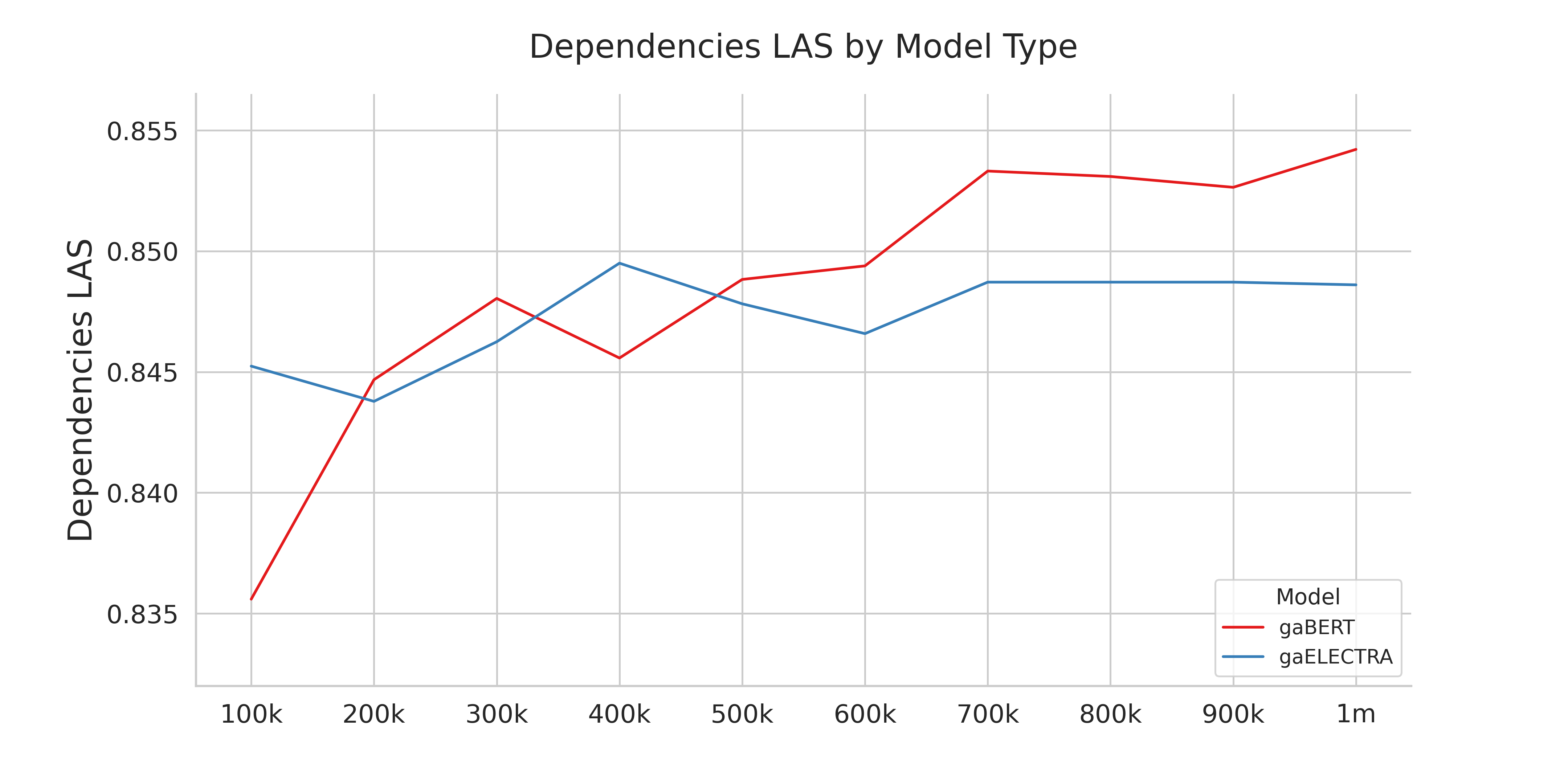}
    \caption{Dependency parsing LAS for each model type.
    Every 100k steps, we show
    the median of five LAS scores obtained from fine-tuning the respective
    model five times with different initialisation.}
    \label{fig:gabert-and-gaelectra}
\end{figure}

Figure~\ref{fig:gabert-and-gaelectra} shows the development LAS of
gaELECTRA and gaBERT for each checkpoint.
The
best gaBERT checkpoint is reached at step 1 million,
which may indicate that there are still gains to be made from training for more steps.
The highest median LAS for gaELECTRA is reached at step 400k.
It is worth noting that although the two models are compared at the same number of steps,
the different pretraining hyperparameters mean they are not trained on the same number of tokens per step.

We also compare the results of the gaELECTRA model
to the other models in Tables~\ref{tab:full-model-results:dev} and
\ref{tab:full-model-results:test}.
gaELECTRA performs slightly below gaBERT but better than both mBERT models and the WikiBERT model.

In terms of the Cloze test experiments:
First, for the original masked token prediction
(Table~\ref{tab:cloze-test-original-token-results}),
gaELECTRA predicted the correct token 75 times,
which is the same number as gaBERT and is slightly below mBERT with continued pretraining, which has a score of 78.
Second,
for the manual evaluation of the tokens generated by each model (Table~\ref{tab:cloze-test-results}),
gaELECTRA predicted
82 matches,
8 mismatches,
1 copy,
and 9 gibberish tokens;
compared to 83, 14, 2 and 1 predicted by gaBERT, respectively.

\section{XLM-R Baseline}\label{sec:roberta} 
\label{sec:xlmr}  

We add another off-the-shelf 
baseline by fine-tuning XLM-R\textsubscript{BASE},
which is a multilingual RoBERTa model introduced by \newcite{conneau-etal-2020-unsupervised}, in the task of multitask dependency parsing and POS and morphological features tagging. 
This model performs better than both variants of mBERT as well as the WikiBERT model but underperforms our two monolingual models, gaBERT and gaELECTRA.

\section{Full Model Results}
\label{sec:full-model-results}

This section examines the results produced by each of our models in more detail and also presents the scores of the additional models we examine,
namely XLM-R\textsubscript{BASE} and gaELECTRA.
\footnote{We tried training a RoBERTa\textsubscript{BASE} model on our data but could not obtain satisfactory LAS scores
(a fine-tuned model achieved a dev LAS of 81.8, which is comparable to mBERT)
and leave finding suitable hyperparameters for this architecture to future work.}
Tables~\ref{tab:full-model-results:dev} and
\ref{tab:full-model-results:test}
list the accuracies for predicting
universal part of speech (UPOS),
treebank-specific part of speech (XPOS)
and morphological features,
as well as
the unlabelled and labelled attachment score (UAS and LAS, respectively) for all models discussed in this paper.

For the multilingual models,
mBERT performs worse than XLM-R\textsubscript{BASE},
which is a strong multilingual baseline.
The monolingual WikiBERT model performs slightly better than mBERT in terms of LAS but is worse than XLM-R\textsubscript{BASE}.
The continued pretraining of mBERT on our data enables us to close the gap between mBERT and XLM-R\textsubscript{BASE}.
gaBERT is still the strongest model for all metrics in terms of test set scores.
gaELECTRA performs slightly below that of gaBERT but better than XLM-R\textsubscript{BASE}.
It should be noted that each row selects the model based on median LAS,
therefore,
all other metrics are those that this selected model achieved.

\begin{table*}
  \centering
  \begin{tabular}{lr|rrrrr}
    \toprule
    {\bf Model} & {\bf UD} & {\bf UPOS} & {\bf XPOS} & {\bf FEATS} & {\bf UAS} & {\bf LAS} \\
      \midrule
      mbert-os     & 2.8 &         95.7  &         94.7  &         89.2  &	       86.9  &         81.8 \\
      xlmr-base-os & 2.8 &         96.4  &         95.1  &         90.6  &	       88.3  &         84.0 \\
      wikibert-os  & 2.8 &         95.9  &         94.9  &         89.4  &         86.8  &         81.9 \\
      mbert-cp     & 2.8 &         97.2  &         95.8  &         92.3  &         88.1  &         84.3 \\
      gabert       & 2.8 &         97.1  & \textbf{96.2} & \textbf{93.1} & \textbf{89.2} & \textbf{85.6} \\
      gaelectra    & 2.8 & \textbf{97.3} &         96.1  &         92.8  &         89.1  &         85.3 \\
      \bottomrule
  \end{tabular}
  \caption{Full model results on development data. For model name abbreviations, see test result table.} 
  \label{tab:full-model-results:dev}
\end{table*}

\begin{table*}
  \centering
  \begin{tabular}{lr|rrrrr}
    \toprule
    {\bf Model} & {\bf UD} & {\bf UPOS} & {\bf XPOS} & {\bf FEATS} & {\bf UAS} & {\bf LAS} \\
      \midrule
      mbert-os     & 2.8 &         95.4  &         94.3  &	       88.6  &         86.2  &	       80.3 \\
      xlmr-base-os & 2.8 &         96.1  &         95.1  &	       90.0  &         87.7  &         82.5 \\
      wikibert-os  & 2.8 &         95.7  &	       94.4  &         88.3  &         85.9  &         80.4 \\
      mbert-cp     & 2.8 &         96.7  &         95.5  &         91.7  &         87.1  &         82.3 \\
      gabert       & 2.8 & \textbf{97.0} & \textbf{95.7} & \textbf{91.8} & \textbf{88.4} & \textbf{84.0} \\
      gaelectra    & 2.8 &         96.9  &         95.5  &         91.5  &         87.6  &         83.1 \\
      \bottomrule
  \end{tabular}
  \caption{Full model results on test data (os = fine-tuned off-the-shelf model, cp = continued pre-training before fine-tuning).} 
  \label{tab:full-model-results:test}
\end{table*}

\end{document}